\documentclass[11pt]{article}
\usepackage{eamt23}
\usepackage{times}
\usepackage{url}
\usepackage{latexsym}
\usepackage[small,bf]{caption} % added MLF 20171211
\usepackage{float}
\usepackage{subfiles}
\usepackage{booktabs}
\usepackage{multirow}
\usepackage[most]{tcolorbox}
\usepackage{comment}
\usepackage{graphicx}
\usepackage{subfig}
\usepackage{hyphenat}

\newcommand*{\myfont}{\fontfamily{lmss}\selectfont}
\DeclareTextFontCommand{\textmyfont}{\myfont}
\newcommand*{\fit}{\fontsize{8pt}{11pt}\selectfont}

\title{Adaptive Machine Translation with Large Language Models}

\author{Yasmin Moslem\\
 \fit{ADAPT Centre} \\
 \fit{School of Computing} \\
 \fit{Dublin City University} \\
 \fit{Dublin, Ireland} \\
 \textrm{\scriptsize{yasmin.moslem@adaptcentre.ie}}  \And
 Rejwanul Haque \\
 \fit{ADAPT Centre} \\
 \fit{Department of Computing} \\
 \fit{South East Technological University} \\
 \fit{Carlow, Ireland} \\
 \textrm{\scriptsize{rejwanul.haque@adaptcentre.ie}} \\\And
 John D. Kelleher \\
 \fit{ADAPT Centre} \\
 \fit{School of Computer Science} \\
 \fit{Technological University Dublin} \\
 \fit{Dublin, Ireland} \\
 \textrm{\scriptsize{john.kelleher@adaptcentre.ie}} \\\And
 Andy Way \\
 \fit{ADAPT Centre} \\
 \fit{School of Computing} \\
 \fit{Dublin City University} \\
 \fit{Dublin, Ireland} \\
 \textrm{\scriptsize{andy.way@adaptcentre.ie}}
 }

\date{}

\begin{document}
\maketitle
\begin{abstract}
\fontsize{10.5pt}{13pt}\selectfont
\nohyphens{
Consistency is a key requirement of high-quality \mbox{translation}. It is especially important to adhere to pre-approved terminology and adapt to corrected translations in domain-specific projects. Machine translation (MT) has achieved significant progress in the area of domain adaptation. However, \mbox{real-time} adaptation remains challenging. Large-scale language models (LLMs) have recently shown interesting capabilities of in-context learning, where they learn to replicate certain input-output text generation patterns, without further fine-tuning. By feeding an LLM at inference time with a prompt that consists of a list of translation pairs, it can then simulate the domain and style characteristics. This work aims to investigate how we can utilize in-context learning to improve real-time adaptive MT. Our extensive experiments show promising results at translation time. For example, LLMs can adapt to a set of in-domain sentence pairs and/or terminology while translating a new sentence. We observe that the translation quality with few-shot in-context learning can surpass that of strong encoder-decoder MT systems, especially for high-resource languages. Moreover, we investigate whether we can combine MT from strong encoder-decoder models with fuzzy matches, which can further improve translation quality, especially for less supported languages. We conduct our experiments across five diverse language pairs, namely English-to-Arabic (EN-AR), English-to-Chinese (EN-ZH), English-to-French (EN-FR), English-to-Kinyarwanda (EN-RW), and English-to-Spanish (EN-ES).
}
\end{abstract}

\begin{figure}[htp]
\captionsetup{font=scriptsize,labelfont=scriptsize}
    \centering
    \includegraphics[width=8.5cm]{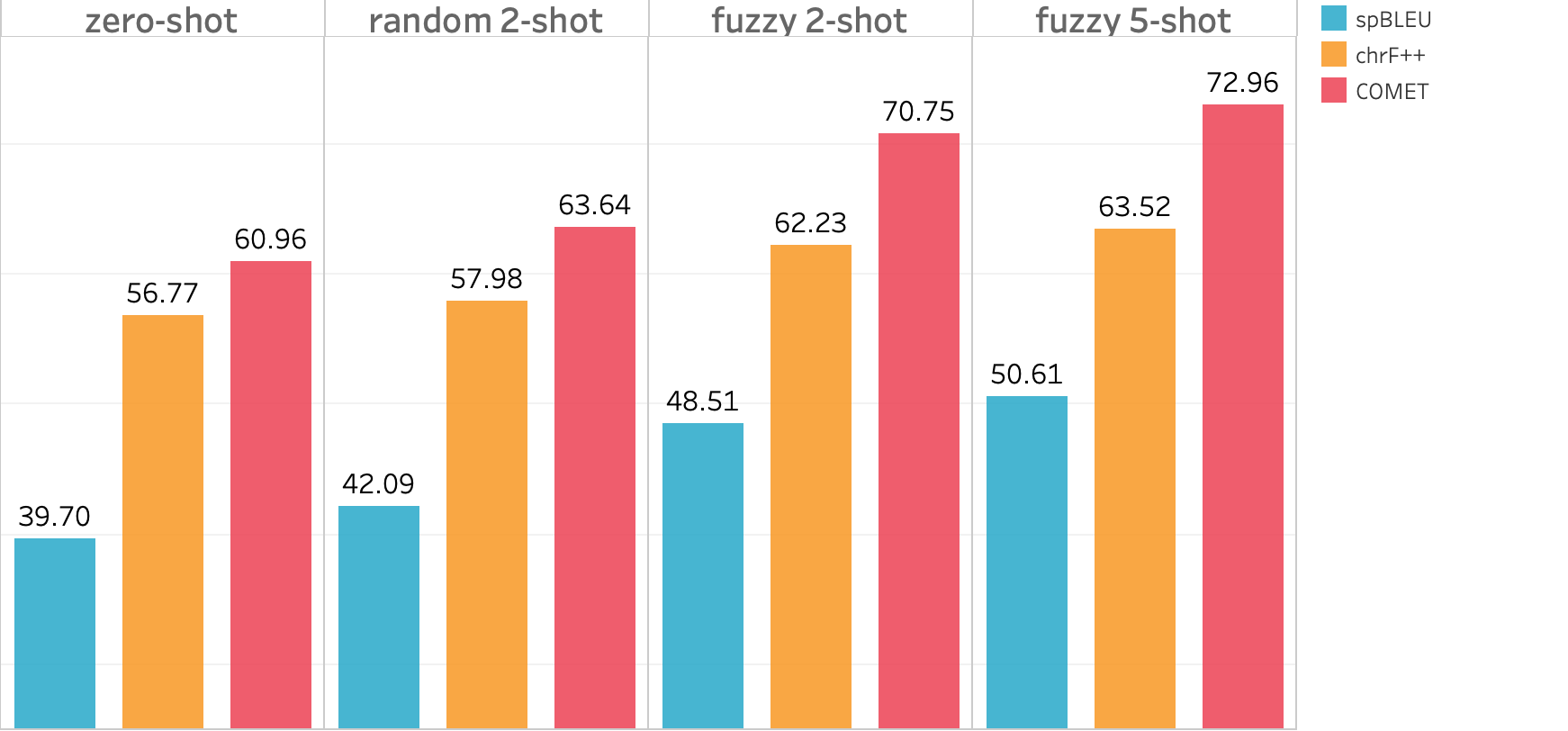}
    \caption{Evaluation results for GPT-3.5 zero-shot, and few-shot translation with random context or fuzzy matches. Average scores across EN-AR, EN-ES, EN-FR, and EN-ZH language pairs. While using a random context outperforms zero-shot translation, using fuzzy matches reveals the best results.}
    \label{fig:context-avg}
\end{figure}

\section{Introduction}

Adaptive MT is a type of machine translation that utilizes feedback from users to improve the quality of the translations over time. Feedback usually includes corrections to previous translations, terminology and style guides, as well as ratings of the quality of the translations. This can be particularly useful for domain-specific scenarios, where baseline MT systems may have insufficient relevant data to accurately translate certain terms or phrases. There are still several challenges to effectively incorporate user feedback into the translation process, especially at inference time. In this work, we use a relatively wide definition of adaptive MT to refer to learning from similar translations (fuzzy matches) found in approved translation memories (TMs) on the fly  \cite{Farajian2017-AdaptiveMT,Wuebker2018-Personalized,Peris2019-OnlineLearning,Etchegoyhen2021-OnlineLearning}, as well as real-time terminology-constrained MT \cite{Hokamp2017-ConstrainedDecoding,Post2018-FastConstrainedDecoding,Dinu2019-TerminologyConstraintsTraining,Michon2020-Terminology}.

Autoregressive decoder-only LLMs, such as GPT-3 \cite{Brown2020-GPT-3,Ouyang2022-InstructGPT}, BLOOM \cite{BLOOM2022}, PaLM \cite{Chowdhery2022-PaLM}, and LLaMA \cite{Touvron2023-LLaMA} are trained to predict the next word given the previous context. During unsupervised pre-training, a language model develops a broad set of pattern recognition abilities. It then uses these abilities at inference time to rapidly recognize and adapt to the desired task. In their experiments, Brown et al.~\shortcite{Brown2020-GPT-3} use the term ``in-context learning" to describe a scenario where a pre-trained language model at inference time learns to replicate certain input-output text generation patterns without further fine-tuning. They show that autoregressive LLMs such as GPT-3 can perform well on diverse tasks, through zero-shot, one-shot, and few-shot in-context learning without weight updates. Instead of asking the model to directly perform a given task, the input can be augmented with relevant examples, which help the model adapt its output. The key idea of in-context learning is to learn from analogy. The model is expected to learn the pattern hidden in the demonstration and accordingly make better predictions \cite{Dong2022-In-contextLearning}.

Previous researchers investigated using neural language models for MT through few-shot in-context learning \cite{Vilar2022-PaLM-Translation} and even in zero-shot settings \cite{Wang2021-LM4MT}. Other researchers proposed using LLMs for generating synthetic domain-specific data for MT domain adaptation \cite{Moslem2022-MT-LM}. Recently, researchers \cite{Agrawal2022-SelectionMT,Zhang2023-PromptingMT} confirmed the importance of in-context example selection for the quality of MT with LLMs.

The main contribution of this paper is investigating the capabilities of LLMs such as \mbox{GPT-3.5}, GPT-4 (including ChatGPT), and BLOOM for real-time adaptive MT through in-context learning. As illustrated by Figure \ref{fig:context-avg}, such LLMs can achieve better translation quality through adapting its output to adhere to the terminology and style used in previously approved translation pairs. In particular, we would like to understand the quality with which such models can perform the following tasks, without any further training:

\begin{itemize}
    \setlength\itemsep{-2pt}
    \item Adapting new translations to match the terminology and style of previously approved TM fuzzy matches, at inference time;
    \item Matching or outperforming the quality of translations generated by encoder-decoder MT models across a number of languages; 
    \item Fixing translations from stronger encoder-decoder MT systems using fuzzy matches, which is especially useful for low-resource languages; and
    \item Terminology-constrained MT, by first defining terminology in the relevant sentences or dataset, and then forcing new translations to use these terms.
\end{itemize}

\section{Experimental Setup}
\label{sec:setup}

In all our experiments, we use GPT-3.5 \emph{text-davinci-003} model via its official API.\footnote{\url{https://openai.com/api/}} For parameters, we use \mbox{\emph{top-p} 1}, with \emph{temperature} 0.3 for the three translation tasks, and 0 for the terminology extraction task.\footnote{To avoid over-generation, the option \emph{stop} can be set to [`\textbackslash{n}']. However, if a new line is generated by the model before the translation, this might result in not generating a translation. Alternatively, over-generation can be manually handled.} For the maximum length of tokens, we observe that French and Spanish tokens can be 3–4 times the number of English source words, while other languages can be longer. Hence, we roughly choose a length multiplier value, which we set to 8 for Arabic, 5 for Chinese and Kinyarwanda, and 4 for French and Spanish. We used batch requests with a batch size of 20 segments.\footnote{For higher values of few-shot translation into Arabic using \emph{text-davinci-003}, we had to decrease the batch size to avoid exceeding the tokens-per-minute limit.} Our scripts are publicly \mbox{available.}\footnote{\url{https://github.com/ymoslem/Adaptive-MT-LLM}}

 As we aim to simulate a document-level scenario where translators are required to adhere to a project's or client's TM, we use the domain-specific dataset, TICO-19 \cite{Anastasopoulos2020-TICO-19}, which includes 3070 unique segments. From now on, we will refer to it as the ``context dataset". We focus on a range of languages with diverse scripts and amounts of resources, namely English as the source language, and Arabic, Chinese, French, Kinyarwanda, and Spanish as the \mbox{target} languages.

\begin{table}[ht]
\captionsetup{font=scriptsize,labelfont=scriptsize}
\centering
\scalebox{.65}{
\begin{tabular}{@{}clcccc@{}}
\toprule
\multicolumn{1}{c}{\textbf{Lang}} & \textbf{Context} & \textbf{spBLEU ↑} & \textbf{chrF++ ↑} & \textbf{TER ↓} & \textbf{COMET ↑} \\ \midrule
\multirow{8}{*}{\textbf{EN-AR}} & zero-shot     & 27.6           & 48.36          & 70.6                                   & 41.28    \\
                                & random 2-shot & 28.94          & 49.35          & 70.55          & 43.32          \\
                                & fuzzy 1-shot   & 36.38          & 55.08          & 63.99          & 55.1           \\
                                & fuzzy 2-shot  & 38.41          & 56.57          & 62.31          & 57.36 \\
                                & fuzzy 3-shot  & 39.75          & 57.52 & 61.12          & 59.68          \\
                                & fuzzy 4-shot  & 40.84          & 58.27          & 60.39          & 62.16          \\
                                & fuzzy 5-shot  & 41.33          & 58.64          & 59.95          & 62.65 \\ 
                                & fuzzy 7-shot  & \textbf{41.81} & \textbf{59.1} & \textbf{59.38} & \textbf{64.01} \\ \midrule
\multirow{5}{*}{\textbf{EN-ES}} & zero-shot     & 53.91          & 72.61    & 36.86                                                & 84.0  \\
                                & random 2-shot & 54.78          & 73.12    & 36.09     
                                        & 85.25 \\
                                & fuzzy 2-shot  & 59.64          & 75.83          & 32.56          & 90.37          \\
                                & fuzzy 5-shot  & 61.24          & 76.73          & 31.32          & 91.51          \\
                                & fuzzy 10-shot & \textbf{61.77} & \textbf{77.05} & \textbf{30.9}  & \textbf{92.0}  \\ \midrule
\multirow{8}{*}{\textbf{EN-FR}} & zero-shot     & 44.87          & 65.29          & 50.34                                  & 58.67          \\
                                & random 2-shot & 45.91          & 65.4           & 49.92          & 57.6           \\
                                & fuzzy 1-shot  & 48.39          & 66.58          & 48.18          & 59.49          \\
                                & fuzzy 2-shot  & 49.79          & 67.41          & 46.79          & 61.38          \\
                                & fuzzy 3-shot  & 50.96          & 68.06          & 45.85          & 61.97          \\
                                & fuzzy 4-shot  & 51.89          & 68.5           & 44.94          & 62.7           \\
                                & fuzzy 5-shot  & 51.94          & 68.43          & 45.09          & 62.81          \\
                                & fuzzy 10-shot & \textbf{53.72} & \textbf{69.39} & \textbf{43.82} & \textbf{63.57} \\ \midrule
\multirow{5}{*}{\textbf{EN-RW}} & zero-shot     & 2.82           & 22.53          & 143.12                                         & N/A            \\
                                & random 2-shot & 3.8            & 25.19          & 129.88         & N/A            \\
                                & fuzzy 2-shot  & 12.23          & 36.66          & 105.54         & N/A            \\
                                & fuzzy 5-shot  & 14.96          & 39.84          & 100.11         & N/A            \\
                                & fuzzy 10-shot & \textbf{17.87} & \textbf{41.44} & \textbf{92.84} & N/A            \\ \midrule
\multirow{5}{*}{\textbf{EN-ZH}} & zero-shot  & 32.41    & 40.82          & 99.45                                           & 59.87          \\
                                & random 2-shot & 38.72  & 44.06      & 87.56  &   68.39            \\
                                & fuzzy 2-shot  & 46.18          & 49.12          & 69.0           & 73.9           \\
                                & fuzzy 5-shot  & 47.94          & 50.28          & 64.96          & 74.86          \\
                                & fuzzy 10-shot & \textbf{49.11} & \textbf{51.22} & \textbf{63.14} & \textbf{75.3}  \\ \bottomrule
\end{tabular}
}
\caption{Adaptive MT with fuzzy matches for GPT-3.5 few-shot in-context learning outperforms using random sentence pairs as context examples. Increasing the number of fuzzy matches can improve the translation quality further. The table shows consistent results for EN-AR, EN-ES, EN-FR, EN-RW, and EN-ZH language pairs.}
\label{tab:fuzzy-context}
\end{table}

\section{Adaptive MT with Fuzzy Matches}
\label{sec:adaptive-MT}

In translation environments, similar approved translated segments are usually referred to as ``fuzzy matches'', and are stored in parallel datasets, known as translation memories (TMs).\footnote{Segments stored in a TM can be smaller than a full sentence (e.g. a title) or larger. However, as most segments in a TM are supposed to be sentence pairs, we use the two words interchangeably throughout the paper.} Researchers have investigated the possibilities of improving MT quality and consistency with fuzzy matches \cite{Knowles2018-Fuzzy,Bulte2019-fuzzy,Xu2020-fuzzy}. Incorporating fuzzy matches into the MT process can help the system generate more accurate translations, and try to ensure adherence to pre-approved terminology and preferred style requirements.

In this set of experiments, we investigate the possibility of forcing the translation of a new sentence pair to adapt to fuzzy matches in the \mbox{context} dataset. To extract fuzzy matches, we use embedding similarity-based retrieval. Previous researchers have shown that approaches that depend on embeddings to retrieve fuzzy matches can outperform those that use Edit Distance \cite{Hosseini2020-DeepMatch,Pham2020-Priming}. To this end, we employ the paraphrase mining module from the Sentence-Transformers library \cite{Reimers2019-SentenceTransformers}. We use the \mbox{\emph{all-MiniLM-L6-v2}} model because of its high accuracy and efficiency.\footnote{\url{https://www.sbert.net/}} For each sentence, we retrieve up to $top\_k$ other sentences. We experiment with diverse values of \mbox{1 to 10} sentence(s) from the \mbox{context} dataset.\footnote{For Arabic, we could only integrate up to 7 matches (not 10 matches) because the tokenizer used by GPT-3.5 generates many more tokens for some Unicode languages, which can easily hit the max length of 4097 tokens. We observe that the issue has been alleviated by newer models.} \mbox{Table} \ref{tab:fuzzy-stats} elaborates on the statistics of fuzzy matches based on their similarity to the new source sentence in 2-shot and 5-shot scenarios.\footnote{While creating prompts, we arrange fuzzy matches in descending order, making higher matches closer to the segment to be translated. We experimented with reversing the order, and there was no significant difference in terms of translation quality.}

The following illustrations show the difference between zero-shot and few-shot translation prompts. In the zero-shot prompt, only the source sentence and language names are provided, encouraging the model to generate the translation. The few-shot prompt incorporates translation examples to influence the style of the output.
%\mbox{Appendix} \ref{a-prompts} provides more examples of the prompts we used throughout our experiments.

\begin{tcolorbox}[enhanced,attach boxed title to top left={yshift=-3mm,yshifttext=-1mm,xshift=3mm},
  colback=blue!1!white,colframe=cyan!90!black,colbacktitle=cyan!90!black,boxrule=1pt,
  left=3pt, top=0pt, width=210pt,center,
  title=Prompt: EN-AR zero-shot translation,fonttitle=\small,
  boxed title style={size=small,colframe=cyan!90!black} ]
  \begin{scriptsize}
  \textmyfont{\emph{
    \begin{itemize}
    \setlength\itemsep{-0.8ex}
    \item[] English: $<$source\_segment$>$
    \item[] Arabic:
\end{itemize}
}}
\end{scriptsize}
\end{tcolorbox}

\begin{tcolorbox}[enhanced,attach boxed title to top left={yshift=-3mm,yshifttext=-1mm,xshift=3mm},
  colback=blue!1!white,colframe=cyan!90!black,colbacktitle=cyan!90!black,boxrule=1pt,
  left=3pt, top=0pt, width=210pt,center,
  title=Prompt: EN-AR two-shot translation,fonttitle=\small,
  boxed title style={size=small,colframe=cyan!90!black} ]
  \begin{scriptsize}
  \textmyfont{\emph{
    \begin{itemize}
    \setlength\itemsep{-0.8ex}
    \item[] English: $<$source\_fuzzy\_match\textsubscript{2}$>$
    \item[] Arabic: $<$target\_fuzzy\_match\textsubscript{2}$>$
    \item[] English: $<$source\_fuzzy\_match\textsubscript{1}$>$
    \item[] Arabic: $<$target\_fuzzy\_match\textsubscript{1}$>$
    \item[] English: $<$source\_segment$>$
    \item[] Arabic:
\end{itemize}
}}
\end{scriptsize}
\end{tcolorbox}

Results illustrated by Figure \ref{fig:context-avg} show that few-shot translation with \mbox{GPT-3.5} using fuzzy matches as context outperforms few-shot translation with random examples, although using random sentence pairs outperforms zero-shot translation. As demonstrated by Table \ref{tab:fuzzy-context}, across five language pairs, adding more fuzzy matches improves translation quality further. At some point, there might be diminishing returns of adding more similar sentences as their similarity score decreases. In other words, increasing the number of fuzzy matches from 2 sentences to 5 or 10 sentences incrementally improves translation quality, but with smaller quality gains.

\begin{figure*}[htp]
\captionsetup{font=scriptsize,labelfont=scriptsize}
    \centering
    \includegraphics[width=17cm]{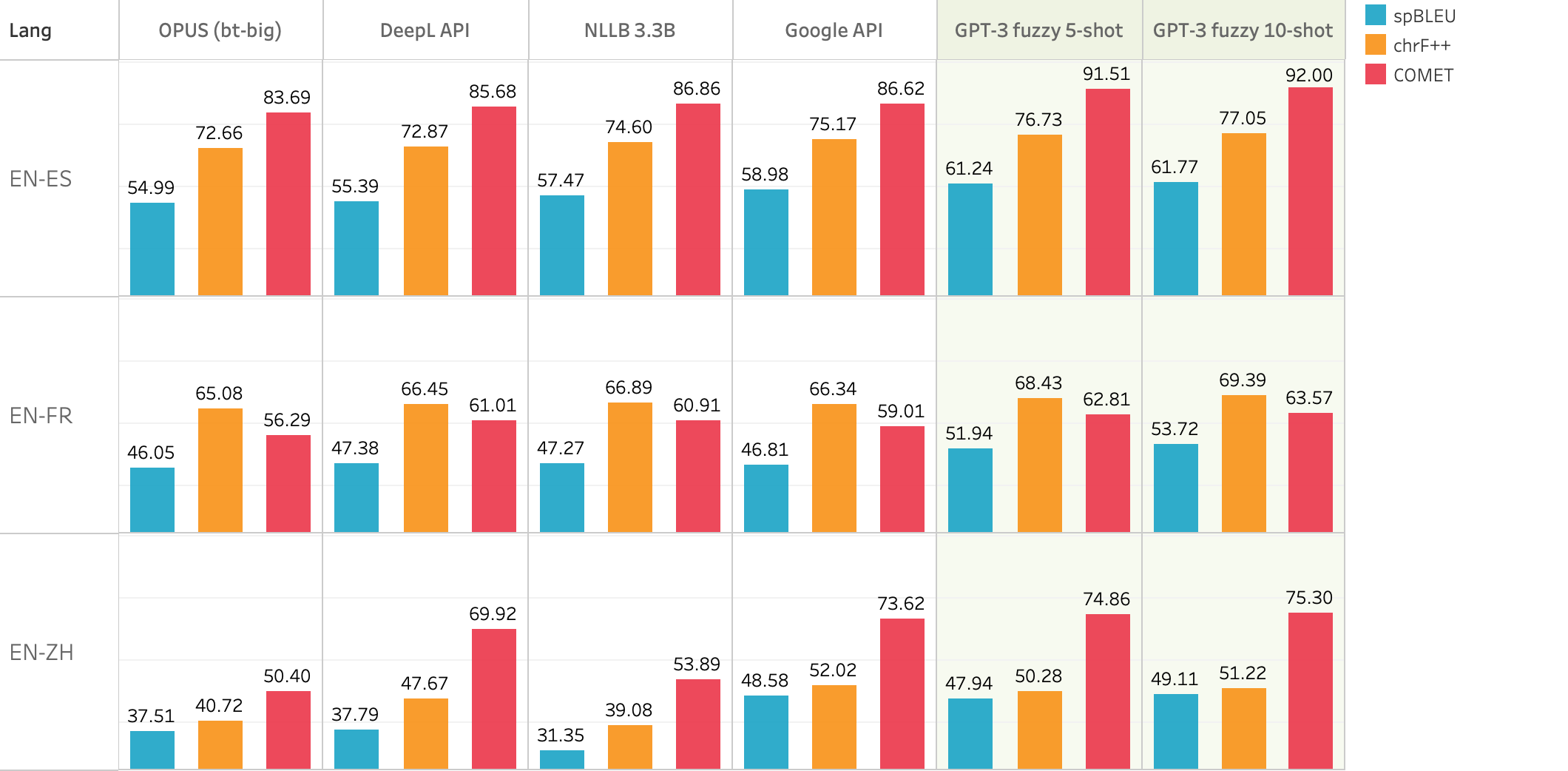}
    \caption{Evaluation results for GPT-3.5 few-shot translation with 5 or 10 fuzzy matches compared to encoder-decoder MT models (DeepL, Google, OPUS, and NLLB). Specifically, for EN-ES, EN-FR, and EN-ZH language pairs, few-shot translation with GPT-3.5 outperforms conventional systems.}
    \label{fig:compare-mt}
\end{figure*}

\begin{table}[H]
\captionsetup{font=scriptsize,labelfont=scriptsize}
\centering
\begin{footnotesize}
\begin{tabular}{@{}lllll@{}}
\toprule
\multicolumn{1}{c}{\multirow{2}{*}{\textbf{\begin{tabular}[c]{@{}c@{}}Similarity\\ Score\end{tabular}}}} &
  \multicolumn{4}{c}{\textbf{Segment Statistics}} \\ \cmidrule(l){2-5} 
\multicolumn{1}{c}{} &
  \multicolumn{2}{c|}{\textbf{fuzzy 2-shot}} &
  \multicolumn{2}{c}{\textbf{fuzzy 5-shot}} \\ \midrule
\textgreater{}90\% & 167   & 2.7\%  & 168   & 1.1\%  \\
89-80\%            & 751   & 12.2\% & 1,103 & 7.2\%  \\
79-70\%            & 1,593 & 25.9\% & 3,143 & 20.5\% \\
69-60\%            & 1,825 & 29.7\% & 4,661 & 30.4\% \\
\textless{}60\%    & 1,804 & 29.4\% & 6,275 & 40.9\% \\ \midrule
Total &
  \multicolumn{2}{l}{6,140 = 3,070*2} &
  \multicolumn{2}{l}{15,350 = 3,070*5} \\ \bottomrule
\end{tabular}
\end{footnotesize}
\caption{Numbers and percentages of segments based on their similarity to the new source segment, in the 2-shot and 5-shot experiments using fuzzy matches for in-context learning. The English source is used to calculate similarity across the 5 language pairs.}
\label{tab:fuzzy-stats}
\end{table}

\section{GPT-3 vs Encoder-Decoder MT Models}
\label{sec:compare-mt}

In this section, we aim to compare evaluation results we obtained from various MT encoder-decoder Transformer-based systems \cite{Vaswani2017-attention} with those from GPT-3.5. To this end, we translate our context dataset with a range of open-source and commercial MT models, including DeepL Translate API,\footnote{DeepL supports French, Spanish and Chinese, but not Arabic and Kinyarwanda.} Google Cloud Translation API, OPUS \cite{Tiedemann2020-Tatoeba},\footnote{We use OPUS models from the Tatoeba-Challenge, specifically the models augmented with back-translation, and trained with Transformer-Big.} and NLLB-200 \cite{NLLB_Team2022}. We converted OPUS and NLLB models to the \mbox{CTranslate2} \cite{Klein2020-Efficient} format with int8 quantization for efficiency. Inference parameters include \emph{beam\_size 4} and \emph{max\_batch\_size 2024}, on a GPU \emph{A100-SXM4-40GB} (Google Colab Pro). For tokenization, we used SentencePiece \cite{Kudo2018-SentencePiece} with the source and target sub-word models provided for each OPUS model, and the multilingual model provided by NLLB for tokenization.\footnote{$flores200\_sacrebleu\_tokenizer\_spm.model$ is used for both tokenization for NLLB and also for spBLEU \cite{Goyal2022-spBLEU} in sacreBLEU.}

We observe that for high-resource languages, adaptive MT with fuzzy matches using \mbox{GPT-3.5} few-shot in-context learning (cf. Section \ref{sec:adaptive-MT}) can outperform strong encoder-decoder MT systems. For the English-to-French and English-to-Spanish language pairs, few-shot translation with \mbox{GPT-3.5} incorporating only 5 fuzzy matches outperforms strong encoder-decoder MT models, as demonstrated by Figure \ref{fig:compare-mt}. For English-to-Chinese translation, only when we used 10 fuzzy matches could we achieve better results. However, for English-to-Arabic and English-to-Kinyarwanda translations, results were not on par with the other three language pairs. The results are detailed in Table \ref{tab:compare-mt}.

% ModernMT
Among the popular adaptive encoder-decoder MT systems is ModernMT.\footnote{\url{https://www.modernmt.com/}} Originally, the system adopted the instance-based adaptation approach proposed by Farajian et al.~\shortcite{Farajian2017-AdaptiveMT}. To control our experiments with ModernMT to match those with \mbox{GPT-3.5} few-shot translation, we created a new TM for each segment to include only the top-10 fuzzy matches for this segment. \mbox{Table}~\ref{tab:compare-mt} illustrates the evaluation results of \mbox{ModernMT} translation with and without a TM. In general, using a TM with ModernMT improves translation quality. Moreover, we observe that zero-shot translation performance (without a TM) of \mbox{ModernMT} outperforms \mbox{GPT-3.5} for the 4 supported language pairs. However, except for English-to-Arabic, few-shot translation with \mbox{GPT-3.5} using either 5 or 10 fuzzy matches outperforms the translation quality of ModernMT using a TM with 10 fuzzy matches per segment, for English-to-Chinese, English-to-French, and English-to-Spanish language pairs.

\section{Incorporating Encoder-Decoder MT}
\label{sec:fix-mt}

As we demonstrated in the previous section, encoder-decoder MT models have achieved high translation quality for several language pairs. \mbox{Nevertheless}, adaptive MT with LLM few-shot in-context learning can surpass such quality, especially for high-resource languages. In this section, we investigate whether we can utilize encoder-decoder MT models to further improve adaptive translation with \mbox{GPT-3.5}. In the next subsections, we study two scenarios:
%\vspace*{-1pt}
\begin{itemize}
    \item appending fuzzy matches with MT from an encoder-decoder model to enhance in-context learning.
    \item translating the source side of fuzzy matches, and using these MT translations for few-shot in-context learning along with the original translations.
\end{itemize}

\newpage
\subsection{Fuzzy matches + new segment MT}
\label{sec:fuzzy-1-mt}

Incorporating a translation from an encoder-decoder MT model with fuzzy matches, we could achieve substantial improvements over the baseline MT performance. As illustrated by Table \ref{tab:fix-mt}, although OPUS English-to-Arabic translation quality outperforms \mbox{GPT-3.5} few-shot translation with 5 fuzzy matches, appending these fuzzy matches with OPUS translation outperforms both OPUS translation only and \mbox{GPT-3.5} translation with fuzzy matches only. Similarly, adding Google English-to-Chinese translation to 5 fuzzy matches outperforms both baselines. Even for the very low-resource English-to-Kinyarwanda language pair, we relatively notice a similar behaviour, using MT outputs of OPUS or NLLB models.

%\begin{figure}[htp]
%\captionsetup{font=scriptsize,labelfont=scriptsize}
%    \centering
%    \includegraphics[width=8.5cm]{img/fix-mt-enar.png}
%    \caption{The English-to-Arabic translation by OPUS is better than the \mbox{GPT-3.5} few-shot translation. When we incorporate both fuzzy matches and OPUS translation for the new segment into \mbox{GPT-3.5} few-shot in-context learning, the generated translation outperforms both baseline translations.}
%    \label{fig:fix-mt-enar}
%\end{figure}

However, we observe that if the translation with only fuzzy matches is significantly better than the encoder-decoder MT baseline, we may not achieve further gains. For example, the \mbox{GPT-3.5} translations with 5 fuzzy matches are already much better than the OPUS translation for English-to-French or Google translation for English-to-Spanish. That is why incorporating the MT output from OPUS or Google did not enhance the \mbox{GPT-3.5} translation quality for these language pairs.

\begin{table}[ht]
\captionsetup{font=scriptsize,labelfont=scriptsize}
\centering
\scalebox{.61}{
\begin{tabular}{@{}clcccc@{}}
\toprule
\textbf{Lang}    & \textbf{System}     & \textbf{spBLEU ↑} & \textbf{chrF++ ↑} & \textbf{TER ↓} & \textbf{COMET ↑} \\ \midrule
\multirow{10}{*}{\textbf{EN-AR}} & OPUS (bt-big) & 43.11    & 60.79    & 57.24    & 63.64      \\
                                 & NLLB 600M     & 35.66    & 54.6     & 62.07    & 54.53      \\
                                 & NLLB 1.2B     & 41.1     & 58.51    & 57.15    & 63.85      \\
                                 & NLLB 3.3B     & 43.42    & 60.11    & 55.58    & 66.8 \\
                                 & Google API    & 43.56    & 61.58    & 57.79    & 65.5    \\
                                 & ModernMT (no TM)    & 47.17    & 62.82    & 53.53    & 66.64   \\
                                 & ModernMT (TM) & \textbf{50.33} & \textbf{65.19}    & \textbf{50.19}    & \textbf{71.0}    \\
                                 & GPT-3 zero-shot     & 27.6     & 48.36    & 70.6     & 41.28    \\
                                 & GPT-3 fuzzy 5-shot  & 41.33    & 58.64    & 59.95    & 62.65    \\ 
                                 & GPT-3 fuzzy 7-shot  & 41.81 & 59.1   & 59.38    & 64.01 \\ \midrule
\multirow{10}{*}{\textbf{EN-ES}} & OPUS (bt-big) & 54.99    & 72.66    & 36.26    & 83.69      \\
                                 & NLLB 600M     & 53.31    & 72.19    & 37.13    & 83.09      \\
                                 & NLLB 1.2B     & 56.1     & 73.85    & 34.96    & 85.91      \\
                                 & NLLB 3.3B     & 57.47    & 74.6     & 33.99    & 86.86      \\
                                 & DeepL API     & 55.39    & 72.87    & 36.21    & 85.68      \\
                                 & Google API    & 58.98    & 75.17    & 32.46    & 86.62      \\
                                 & ModernMT (no TM)    & 57.09    &  74.2    &  34.27   &  85.53  \\
                                 & ModernMT (TM)       &  59.22   &  75.4    &  32.79   &  86.99  \\
                                 & GPT-3 zero-shot     & 53.91    & 72.61    & 36.86    & 84.0     \\
                                 & GPT-3 fuzzy 5-shot  & 61.24    & 76.73    & 31.32    & 91.51    \\
                                 & GPT-3 fuzzy 10-shot & \textbf{61.77} & \textbf{77.05} &
                                 \textbf{30.9}  & \textbf{92.0}   \\ \midrule
\multirow{11}{*}{\textbf{EN-FR}} & OPUS (bt-big)  & 46.05    & 65.08    & 49.8     & 56.29          \\
                                 & NLLB 600M     & 43.25    & 64.17    & 51.28    & 56.16          \\
                                 & NLLB 1.2B     & 46.3     & 66.25    & 48.68    & 59.76          \\
                                 & NLLB 3.3B     & 47.27    & 66.89    & 48.19    & 60.91          \\
                                 & DeepL API     & 47.38    & 66.45    & 48.47    & 61.01          \\
                                 & Google API    & 46.81    & 66.34    & 47.01    & 59.01          \\
                                 & ModernMT (no TM)  &   47.17  &   66.28  &   47.91  &   58.46    \\
                                 & ModernMT (TM)     &   49.24  &   67.41  &   46.17  &   59.84    \\
                                 & GPT-3 zero-shot     & 44.87    & 65.29    & 50.34    & 58.67    \\
                                 & GPT-3 fuzzy 5-shot  & 51.94    & 68.43    & 45.09    & 62.81    \\
                                 & GPT-3 fuzzy 10-shot & \textbf{53.72}    & \textbf{69.39}    & \textbf{43.82} & \textbf{63.57}   \\ \midrule
\multirow{8}{*}{\textbf{EN-RW}} & OPUS (Tatoeba 2021) & 1.38     & 15.32    & 153.58    & N/A      \\
                                 & OPUS (2020)   & 5.58     & 27.05    & 101.25   & N/A            \\
                                 & NLLB 600M     & 19.46    & 47.61    & 80.01    & N/A            \\
                                 & NLLB 1.2B     & 23.6     & 50.73    & 74.53    & N/A            \\
                                 & NLLB 3.3B     & \textbf{25.17} & \textbf{52.59} & \textbf{73.06} & N/A  \\
                                 & Google API    & 20.63    & 48.37    & 73.54    & N/A     \\
                                 & GPT-3 zero-shot     & 2.82     & 22.53      & 143.12    & N/A      \\
                                 & GPT-3 fuzzy 5-shot  & 14.96    & 39.84    & 100.11      & N/A     \\      
                                 & GPT-3 fuzzy 10-shot & 17.87    & 41.44    & 92.84 & N/A   \\ \midrule
\multirow{9}{*}{\textbf{EN-ZH}}  & OPUS (bt-big) & 37.51    & 40.72    & 121.49      & 50.4        \\
                                 & NLLB 600M     & 24.9     & 33.87    & 109.37      & 39.28       \\
                                 & NLLB 1.2B     & 29.02    & 37.45    & 110.22      & 50.05       \\
                                 & NLLB 3.3B     & 31.35    & 39.08    & 109.52      & 53.89       \\
                                 & DeepL API     & 37.79    & 47.67    & 100.83      & 69.92       \\
                                 & Google API    & 48.58    & \textbf{52.02} & 70.87    & 73.62    \\
                                 & ModernMT (no TM)    &  37.61   &  48.46   &  102.18  &  67.45   \\
                                 & ModernMT (TM)       &  39.85   &  50.95   &  101.53  &  69.64   \\
                                 & GPT-3 zero-shot     & 32.41    & 40.82    & 99.45    & 59.87    \\
                                 & GPT-3 fuzzy 5-shot  & 47.94    & 50.28    & 64.96    & 74.86    \\
                                 & GPT-3 fuzzy 10-shot & \textbf{49.11} & 51.22    & \textbf{63.14} & \textbf{75.3} \\ \bottomrule
\end{tabular}
}
\caption{Comparing GPT-3.5 few-shot translation using fuzzy matches with encoder-decoder MT systems, DeepL Translate API, Google Cloud Translation API, OPUS (Tatoeba-Challenge, with back-translation and Transformer-Big), and NLLB-200 (600M, 1.2B \& 3.3B parameters).}
\label{tab:compare-mt}
\end{table}

\subsection{Fuzzy matches + all segments MT}
\label{sec:fuzzy-all-mt}

In Section \ref{sec:fuzzy-1-mt}, we added MT of the new segment from an encoder-decoder model to fuzzy matches, which enhanced \mbox{GPT-3.5} in-context learning. In this experiment, we include MT for all fuzzy matches and also for the new source segment to be translated. For the English-to-Kinyarwanda and English-to-Spanish language pairs, it is not clear whether including MT for all in-context examples can significantly outperform including MT for only the new source segment to be translated. Again, this depends on the quality of the original MT and requires further investigation.
%\enlargethispage{0.5\baselineskip}

\section{Bilingual Terminology Extraction}
\label{sec:term-extract}

Terminology extraction is the task of automatically defining domain-specific terms in a dataset. Extracted terms are naturally used for building glossaries to help translators. Furthermore, it is possible to improve MT performance through finding sentences that include these terms and fine-tuning the system with them \cite{Hu2019-LexiconInduction,Haque2020-Terminology}. 

In this set of experiments, we ask \mbox{GPT-3.5} to extract 5 bilingual terms from each sentence pair in the context dataset. For parameters, we use temperature 0 and $top\_p$ 1.

\begin{table}[H]
\captionsetup{font=scriptsize,labelfont=scriptsize}
\centering
\begin{footnotesize}
\begin{tabular}{@{}ccccc@{}}
\toprule
\textbf{Lang}  & \textbf{Sentences} & \textbf{Terms} & \textbf{Correct} & \textbf{\%} \\ \midrule
\textbf{EN-AR} & 500                & 2,500          & 2,427            & 97.08         \\
\textbf{EN-ES} & 500                & 2,500          & 2,397            & 95.88          \\
\textbf{EN-FR} & 500                & 2,500          & 2,382            & 95.28           \\ \bottomrule
\end{tabular}
\end{footnotesize}
\caption{Human evaluation results for the terminology extraction task for English-to-Arabic (EN-AR), English-to-Spanish (EN-ES), and English-to-French (EN-FR) language pairs. The majority of the terms that GPT-3 extracted ($>$\,95\%) were accurate.}
\label{tab:term-extract}
\end{table}

Human evaluation was performed for Arabic, French,\footnote{We observe that the original English-to-French TICO-19 dataset includes several misaligned translation pairs. This can negatively affect the quality of tasks using such sentences. That is why it is important to filter parallel datasets to remove possible misalignments. The evaluation sample has been manually refined to include only well-aligned translation pairs. Automatic semantic filtering approaches can be applied to large datasets.} and Spanish. We provided the evaluators with a random sample of 500 sentences and their extracted terms. They were asked to use a 0-1 scale to determine whether each source and target term were equivalent, and whether the extracted terms were actually in the sentence pair (relevant inflexions are acceptable). In several cases where the evaluators marked the extracted term pair with 0, the model had made up either the source, target, or both; although it might be correct, it was not in the provided sentence pair. In other cases, the extracted term was partial, sometimes due to reaching the maximum length of tokens. Nevertheless, as Table \ref{tab:term-extract} illustrates, the majority of the terms in the provided sample were accurately extracted by the model.

\begin{table*}[ht]
\captionsetup{font=scriptsize,labelfont=scriptsize}
\centering
\begin{scriptsize}
\begin{tabular}{@{}clcccc@{}}
\toprule
\textbf{Lang} & \multicolumn{1}{c}{\textbf{System}} & \textbf{spBLEU ↑} & \textbf{chrF++ ↑} & \textbf{TER ↓} & \textbf{COMET ↑} \\ \midrule
\multirow{3}{*}{\textbf{EN-AR}} & MT (OPUS)                     & 43.11          & 60.79          & 57.24                                     & 63.64          \\
                                & GPT-3 fuzzy 5-shot            & 41.33          & 58.64          & 59.95     & 62.65          \\
                                & GPT-3 fuzzy 5-shot + 1-MT     & \textbf{45.9}  & \textbf{62.9}  & \textbf{55.14} & \textbf{67.74}   \\ \midrule
\multirow{7}{*}{\textbf{EN-ES}} & MT (Google)                   & 58.98          & 75.17          & 32.46                                     & 86.62          \\
                                & GPT-3 fuzzy 2-shot            & 59.64          & 75.83          & 32.56     & 90.37          \\
                                & GPT-3 fuzzy 2-shot + 1-MT     & 59.82          & 75.73          & \textbf{32.16} & 89.0           \\
                                & GPT-3 fuzzy 2-shot + all-MT   & \textbf{60.2}  & \textbf{76.06} & 32.32     & \textbf{92.0}  \\ \cmidrule(l){2-6} 
                                & GPT-3 fuzzy 5-shot            & \textbf{61.24} & \textbf{76.73} & \textbf{31.32} & 91.51          \\
                                & GPT-3 fuzzy 5-shot + 1-MT     & 60.49          & 76.16          & 31.49     & 89.55          \\
                                & GPT-3 fuzzy 5-shot + all-MT   & 61.1           & 76.52          & 31.8      & \textbf{92.07} \\ \midrule
\multirow{3}{*}{\textbf{EN-FR}} & MT (OPUS)                     & 46.05          & 65.08          & 49.8                                      & 56.29          \\
                                & GPT-3 fuzzy 5-shot            & \textbf{51.94} & \textbf{68.43} & \textbf{45.09} & \textbf{62.81} \\
                                & GPT-3 fuzzy 5-shot + 1-MT     & 47.95          & 66.72          & 48.34     & 59.69          \\ \midrule
\multirow{7}{*}{\textbf{EN-RW}} & MT \#1 (Google) & 20.63 & 48.37  & 73.54  & N/A     \\
                                & GPT-3 fuzzy 5-shot            & 14.96          & 39.84          & 100.11    & N/A            \\
                                & GPT-3 fuzzy 5-shot + 1-MT \#1 & 22.51 & \textbf{49.69} & \textbf{72.97} & N/A   \\ 
                                & GPT-3 fuzzy 5-shot + all-MT \#1 & \textbf{25.01} & 49.43 & 74.75 & N/A  \\
                                \cmidrule(l){2-6}
                                & MT \#2 (NLLB 3.3B)            & 25.17          & 52.59          & 73.06     & N/A            \\
                                & GPT-3 fuzzy 5-shot + 1-MT \#2 & 25.59 & 53.12 & \textbf{72.73} & N/A     \\
                                & GPT-3 fuzzy 5-shot + all-MT \#2 & \textbf{27.52} & \textbf{53.23} & 73.79 & N/A  \\ \midrule
\multirow{3}{*}{\textbf{EN-ZH}} & MT (Google)                   & 48.58          & 52.02          & 70.87                                     & 73.62          \\
                                & GPT-3 fuzzy 5-shot            & 47.94          & 50.28          & \textbf{64.96} & \textbf{74.86}          \\
                                & GPT-3 fuzzy 5-shot + 1-MT     & \textbf{49.45} & \textbf{52.4}  & 67.81     & 74.61 \\ \bottomrule
\end{tabular}
\end{scriptsize}

\caption{Combining fuzzy matches with high-quality MT from encoder-decoder systems can improve translation quality with GPT-3.5 few-shot in-context learning, especially for low-resource and medium-resource languages. 1-MT refers to appending fuzzy matches with the MT of the segment to be translated, while all-MT refers to additionally adding MT for each segment of the fuzzy matches along with its approved translation. For EN-AR and EN-RW improvements are clearer than for EN-ES, EN-FR and EN-ZH, potentially due to the limited support of EN-AR and EN-RW by GPT-3.5, which made them benefit more from incorporating MT from stronger encoder-decoder models.}
\label{tab:fix-mt}
\end{table*}

%\vspace*{-10pt}

%\begin{figure*}[htp]
%\captionsetup{font=scriptsize,labelfont=scriptsize}
%    \centering
%    \subfloat[\raggedright\tiny Zero-shot translation with max 5 glossary terms found in the source]{{\includegraphics[width=8cm]{img/term-constrain-zero.png}}}
    %\qquad
%    \subfloat[\raggedright\tiny Few-shot translation with both fuzzy matches and glossary terms]{{\includegraphics[width=8cm]{img/term-constrain-few.png}}}
%    \caption{Terminology-constrained MT with GPT-3. Evaluation results across EN-AR, EN-ES, EN-FR, and EN-ZH language pairs. Integration of terms from \mbox{a pre-approved} glossary improves translation for both zero-shot and 2-shot scenarios, although gains from zero-shot prediction are significantly higher.}
%    \label{fig:term-constrain-zero}
%\end{figure*}

\section{Terminology-Constrained MT}
\label{sec:term-constrain}

As observed in Section \ref{sec:adaptive-MT}, adding more fuzzy matches enhances in-context learning and hence improves translation quality. However, early in a real-world translation project, we might not have so many fuzzy matches. By incorporating domain-specific terminology, the system can produce translations that are more accurate and consistent with the terminology used in that field. In this section, we investigate integrating terms in the process when there are $N$ fuzzy matches. For example, if we have only two fuzzy matches, we either extract terms from these similar sentences or from a \mbox{glossary}, and use those that match up to 5-gram phrases in the source sentence to be translated. In this work, we use the terminology extraction process elaborated in Section \ref{sec:term-extract}. Obviously, if a pre-approved glossary is available, it can be used instead. We investigate three scenarios:

\vspace*{-3pt}
\begin{itemize}
    \setlength\itemsep{-2pt}
    \item Few-shot translation with 2 fuzzy matches and their terms. As we do not have terms for the segment to be translated, we use terms from the 2 fuzzy matches if they are found in a set of n-grams (1-5) of the source segment to be translated. Integrating terms into two-shot prediction, i.e. using both terms and two fuzzy matches for in-context learning, \mbox{outperforms} using fuzzy matches only.

    \item We automatically compile a glossary including all terms from the dataset, with 2+ frequency, and up to 5-grams. If there are multiple targets for the same source, the term pair with the highest frequency is selected. Stop words and terms with empty source or target sides are excluded. The list is sorted by n-gram length, so terms with longer n-grams are prioritized.  As illustrated by Table \ref{tab:term-constrain}, integrating terms from a glossary outperforms adding terms from only two fuzzy matches, most likely due to the diversity that this option offers. In prompts (cf. Appendix \ref{a-prompts}), we use terms found in a set of n-grams (1-5) of the source segment to be translated. We experiment with adding maximum 5 terms and maximum 10 terms, which does not show a huge difference in performance; in some cases only a smaller number of terms is available in the glossary.

    \item Zero-shot translation, i.e. without any fuzzy matches. This is similar to the previous scenario, except that we only use terms from the glossary. In zero-shot prediction, adding terms from the glossary improves translation quality. As shown in Table \ref{tab:term-constrain}, improvements are significant across all 5 language pairs.
\end{itemize}

\begin{table*}[ht]
\captionsetup{font=scriptsize,labelfont=scriptsize}
\centering
\begin{scriptsize}
\begin{tabular}{@{}clcccc@{}}
\toprule
\textbf{Lang} & \textbf{GPT-3.5 Context}                                            & \textbf{spBLEU ↑} & \textbf{chrF++ ↑} & \textbf{TER ↓} & \textbf{COMET ↑} \\ \midrule
\multirow{6}{*}{\textbf{EN-AR}} & zero-shot     & 27.6                   & 48.36          &       70.6     &                                  41.28          \\
                                & zero-shot + max 5 terms (glossary)     & \textbf{35.38} & \textbf{54.53} & \textbf{65.36}  & \textbf{54.91} \\ \cmidrule(l){2-6} 
                                & fuzzy 2-shot                           & 38.41          & 56.57          & 62.31           & 57.36          \\
                                & fuzzy 2-shot + terms (fuzzy)           & 39.38          & 57.22          & 62.01           & 59.36          \\
                                & fuzzy 2-shot + max 5 terms (glossary)  & 41.27          & 58.84          & 60.09           & 62.17          \\
              & fuzzy 2-shot + max 10 terms (glossary)                     & \textbf{41.95}    & \textbf{59.34}    & \textbf{59.45} & \textbf{62.48}   \\ \midrule
\multirow{6}{*}{\textbf{EN-ES}} & zero-shot     & 53.91                  & 72.61          & 36.86          &                                  84.0           \\
                                & zero-shot + max 5 terms (glossary)     & \textbf{55.99} & \textbf{74.18} & \textbf{35.3}   & \textbf{87.21} \\ \cmidrule(l){2-6} 
                                & fuzzy 2-shot                           & 59.64          & 75.83          & 32.56           & 90.37          \\
                                & fuzzy 2-shot + terms (fuzzy)           & 59.66          & 75.91          & 32.53           & 90.04          \\
                                & fuzzy 2-shot + max 5 terms (glossary)  & 60.5           & 76.55          & \textbf{31.93}  & \textbf{91.05} \\
                                & fuzzy 2-shot + max 10 terms (glossary) & \textbf{60.54} & \textbf{76.58} & 32.02           & \textbf{91.05} \\ \midrule
\multirow{7}{*}{\textbf{EN-FR}} & zero-shot                              & 44.87          & 65.29          & 50.34           & 58.67          \\
                                & zero-shot + max 5 terms (glossary)     & \textbf{45.94} & \textbf{66.01} & \textbf{49.22}  & \textbf{59.78} \\ \cmidrule(l){2-6} 
                                & fuzzy 2-shot                           & 49.79          & 67.41          & 46.79           & 61.38          \\
                                & fuzzy 2-shot + terms (fuzzy)           & \textbf{50.58}          & \textbf{67.93} & 45.81           & 62.04          \\
                                & fuzzy 2-shot + max 3 terms (glossary)  & 50.46          & 67.69          & 46.22           & \textbf{68.94} \\
                                & fuzzy 2-shot + max 5 terms (glossary)  & 50.55 & 67.78         & \textbf{46.19}  & 60.24           \\
                                & fuzzy 2-shot + max 10 terms (glossary) & 49.64          & 66.86          & 47.34           & 58.57          \\ \midrule
\multirow{7}{*}{\textbf{EN-RW}} & zero-shot     & 2.82                   & 22.53          & 143.12         &                                  N/A            \\
                                & zero-shot + max 5 terms (glossary)     & \textbf{7.26}  & \textbf{30.83} & \textbf{115.44} & N/A            \\ \cmidrule(l){2-6} 
                                & fuzzy 2-shot                           & 12.23          & 36.66          & 105.54          & N/A            \\
                                & fuzzy 2-shot + terms (fuzzy)           & 12.43          & 36.48          & 102.22          & N/A            \\
                                & fuzzy 2-shot + max 5 terms (glossary)  & 15.34          & 39.96          & 96.09           & N/A            \\
                                & fuzzy 2-shot + max 10 terms (glossary) & \textbf{15.49}    & \textbf{40.53}    & \textbf{96.0}  & N/A          \\
                            \midrule
\multirow{7}{*}{\textbf{EN-ZH}} & zero-shot                              & 32.41          & 40.82          & 99.45           & 59.87          \\
                                & zero-shot + max 5 terms (glossary)     & 36.31          & 44.72          & 96.45           & 68.6           \\
              & \multicolumn{1}{l}{zero-shot + max 10 terms (glossary)}    & \textbf{36.64}    & \textbf{45.06}    & \textbf{96.24} & \textbf{68.94}   \\ \cmidrule(l){2-6} 
                                & fuzzy 2-shot                           & 46.18          & 49.12          & 69.0            & \textbf{73.9}  \\
                                & fuzzy 2-shot + terms (fuzzy)           & 46.16          & 49.11          & \textbf{68.79}  & 73.41          \\
                                & fuzzy 2-shot + max 5 terms (glossary)  & \textbf{46.6}  & \textbf{49.51} & 69.46           & 73.88          \\
                                & fuzzy 2-shot + max 10 terms (glossary) & 46.31          & 49.25          & 69.39           & 73.57          \\ \bottomrule
\end{tabular}
\end{scriptsize}
\caption{Terminology-constrained MT with GPT 3.5 outperforms both zero-shot and 2-shot translation with fuzzy matches, although gains are much higher for zero-shot translation. For zero-shot translation, we experimented with adding terms from a glossary. For 2-shot translation with fuzzy matches, we compared adding terms from these 2 fuzzy matches to adding terms from a glossary. The latter revealed better results.}
\label{tab:term-constrain}
\end{table*}

We conducted human evaluation for English-to-Arabic, English-to-French, and English-to-Spanish terminology-constrained MT, to see to what extent the model adheres to the required terms, and how this affects the overall translation quality. The evaluators are professional linguists in the respective languages. We provided the evaluators with 4 sets of 100 randomly selected sentence pairs (zero-shot, zero-shot with glossary terms, fuzzy two-shot, and fuzzy two-shot with glossary terms). They were asked to evaluate the sentence-level translation quality on a 1-4 scale \cite{Coughlin2003-MT-eval} and the usage of each provided term in the translation on a 0-1 scale, as elaborated by \mbox{Table \ref{tab:human-eval}.}

\begin{table}[H]
\captionsetup{font=scriptsize,labelfont=scriptsize}
\centering
\begin{scriptsize}
\begin{tabular}{@{}clcc@{}}
\toprule
\multicolumn{1}{l}{\textbf{\textbf{Lang}}} & \multicolumn{1}{c}{\textbf{GPT-3 Context}} & \textbf{\textbf{Human Eval. ↑}} & \textbf{\textbf{Terms ↑}} \\ \midrule
\multirow{5}{*}{\textbf{EN-AR}} & Zero-shot              & 2.80          & 0.67          \\
                                & Zero-shot + glossary terms      & \textbf{3.19} & \textbf{0.94} \\ \cmidrule(l){2-4}
                                & Fuzzy two-shot         & 2.89          & 0.80          \\
                                & Fuzzy two-shot + glossary terms & \textbf{3.03} & \textbf{0.94} \\ \midrule
\multirow{4}{*}{\textbf{EN-ES}} & Zero-shot              & 3.76          & 0.87          \\
                                & Zero-shot + glossary terms      & \textbf{3.93} & \textbf{0.96} \\ \cmidrule(l){2-4} 
                                & Fuzzy two-shot         & 3.77          & 0.89          \\
                                & Fuzzy two-shot + glossary terms & \textbf{3.84} & \textbf{0.97} \\ \midrule
\multirow{4}{*}{\textbf{EN-FR}} & Zero-shot              & 3.55          & 0.89          \\
                                & Zero-shot + glossary terms      & \textbf{3.64} & \textbf{0.97} \\ \cmidrule(l){2-4} 
                                & Fuzzy two-shot         & 3.50          & 0.91          \\
                                & Fuzzy two-shot + glossary terms & \textbf{3.55} & \textbf{0.92} \\ \bottomrule
\end{tabular}
\end{scriptsize}
\caption{Human evaluation of terminology-constrained MT, for EN-AR, EN-ES, and EN-FR. The results cover zero-shot and two-shot translation without and with (maximum 5) glossary terms. The column ``Human Eval." refers to the average evaluation score on a 1-4 scale. The column ``Terms" refers to the average number of terms that the model has successfully transferred into the translation on a 0-1 scale.}
\label{tab:human-eval}
\end{table}

According to the evaluators, for Arabic, French and Spanish, terminology-constrained MT successfully transferred the provided glossary terms into the target more often than zero-shot and few-shot translation without terminology incorporation. In several cases, forcing glossary terms to be used could help improve the overall translation quality; however, sometimes it was detrimental to grammatical accuracy. Although we provided the model with longer terms before shorter ones, contradictory terms can hurt translation quality. Hence, it might be better to exclude shorter terms if they overlap with longer ones.\footnote{For example, ``New York Times" can be transferred without translation into the target, while ``New York" might be translated. If the model is provided with both terms while it is actually supposed to use the former, this can cause confusion.} In production workflows, linguists can be provided with translation alternatives with and without fuzzy matches and/or terminology to be able to use the best translation. Alternatively, automatic quality estimation can be conducted to select the best translation.

Among interesting observations that human evaluation reveals is that in few-shot translation with fuzzy matches (even \emph{without} terms), the number of successfully used terms is more than those in zero-shot translation. This can help enhance \mbox{consistency} with approved translations. Moreover, incorporating glossary terms in a zero-shot prompt can result in quality gains comparable to those of few-shot translation with fuzzy matches.

\section{ChatGPT}

At the time of writing this paper, OpenAI has released new conversational models, publicly \mbox{referred} to as ChatGPT. This range of models includes: GPT-3.5 Turbo and GPT-4. In this section, we briefly investigate the translation capabilities of these models compared to GPT-3.5 Davinci. Generally, we observe that both of the new models solve some tokenization issues, especially for non-Latin languages such as \mbox{Arabic}. While \emph{gpt-3.5-turbo} is more efficient than \emph{text-davinci-003}, it shows comparable quality for both zero-shot and few-shot translation (with fuzzy matches). The newest model \emph{gpt-4} provides better zero-shot translation quality, while the quality of few-shot translation is relatively similar to that of the two other models. Table \ref{tab:chatgpt} demonstrates the results.

%\vspace*{-4pt}
\begin{table}[H]
\captionsetup{font=scriptsize,labelfont=scriptsize}
\centering
\scalebox{0.56}{
\begin{tabular}{@{}ccccccc@{}}
\toprule
\textbf{Lang} &
  \textbf{Model} &
  \textbf{Context} &
  \textbf{spBLEU ↑} &
  \textbf{chrF++ ↑} &
  \textbf{TER ↓} &
  \textbf{COMET ↑} \\ \midrule
\multirow{6}{*}{\textbf{EN-AR}} &
  GPT-3.5 Davinci &
  \multirow{3}{*}{0-shot} &
  27.6 &
  48.36 &
  70.6 &
  41.28 \\
 &
  GPT-3.5 Turbo &
   &
  38.06 &
  56.35 &
  61.34 &
  62.68 \\
 &
  GPT-4 &
   &
  \textbf{40.29} &
  \textbf{57.86} &
  \textbf{59.55} &
  \textbf{64.25} \\ \cmidrule(l){2-7} 
 &
  GPT-3.5 Davinci &
  \multirow{3}{*}{2-shot} &
  38.41 &
  56.57 &
  62.31 &
  57.36 \\
 &
  GPT-3.5 Turbo &
   &
  46.04 &
  62.18 &
  55.03 &
  73.35 \\
 &
  GPT-4 &
   &
  \textbf{47.52} &
  \textbf{63.28} &
  \textbf{53.04} &
  \textbf{73.7} \\ \midrule
\multirow{6}{*}{\textbf{EN-ES}} &
  GPT-3.5 Davinci &
  \multirow{3}{*}{0-shot} &
  53.91 &
  72.61 &
  36.86 &
  84.0 \\
 &
  GPT-3.5 Turbo &
   &
  52.91 &
  70.87 &
  38.86 &
  82.28 \\
 &
  GPT-4 &
   &
  \textbf{56.93} &
  \textbf{74.41} &
  \textbf{34.35} &
  \textbf{87.89} \\ \cmidrule(l){2-7} 
 &
  GPT-3.5 Davinci &
  \multirow{3}{*}{2-shot} &
  59.64 &
  75.83 &
  32.56 &
  90.37 \\
 &
  GPT-3.5 Turbo &
   &
  \textbf{60.35} &
  \textbf{76.51} &
  32.05 &
  91.57 \\
 &
  GPT-4 &
   &
  60.16 &
  \textbf{76.51} &
  \textbf{31.77} &
  \textbf{91.86} \\ \midrule
\multirow{6}{*}{\textbf{EN-FR}} &
  GPT-3.5 Davinci &
  \multirow{3}{*}{0-shot} &
  44.87 &
  65.29 &
  50.34 &
  58.67 \\
 &
  GPT-3.5 Turbo &
   &
  46.85 &
  66.75 &
  48.31 &
  61.34 \\
 &
  GPT-4 &
   &
  \textbf{47.39} &
  \textbf{67.14} &
  \textbf{48.03} &
  \textbf{61.93} \\ \cmidrule(l){2-7} 
 &
  GPT-3.5 Davinci &
  \multirow{3}{*}{2-shot} &
  49.79 &
  67.41 &
  46.79 &
  61.38 \\
 &
  GPT-3.5 Turbo &
   &
  \textbf{49.88} &
  68.33 &
  46.27 &
  63.62 \\
 &
  GPT-4 &
   &
  49.75 &
  \textbf{68.38} &
  \textbf{45.97} &
  \textbf{64.04} \\ \midrule
\multicolumn{1}{l}{\multirow{6}{*}{\textbf{EN-RW}}} &
  GPT-3.5 Davinci &
  \multirow{3}{*}{0-shot} &
  2.82 &
  22.53 &
  143.12 &
  N/A \\
\multicolumn{1}{l}{} &
  GPT-3.5 Turbo &
   &
  5.31 &
  29.77 &
  114.34 &
  N/A \\
\multicolumn{1}{l}{} &
  GPT-4 &
   &
  \textbf{8.95} &
  \textbf{35.28} &
  \textbf{93.15} &
  N/A \\ \cmidrule(l){2-7} 
\multicolumn{1}{l}{} &
  GPT-3.5 Davinci &
  \multirow{3}{*}{2-shot} &
  12.23 &
  36.66 &
  105.54 &
  N/A \\
\multicolumn{1}{l}{} &
  GPT-3.5 Turbo &
   &
  12.49 &
  39.37 &
  105.51 &
  N/A \\
\multicolumn{1}{l}{} &
  GPT-4 &
   &
  \textbf{16.78} &
  \textbf{44.21} &
  \textbf{83.31} &
  N/A \\ \midrule
\multicolumn{1}{l}{\multirow{6}{*}{\textbf{EN-ZH}}} & GPT-3.5 Davinci & \multirow{3}{*}{0-shot} & 32.41 & 40.82 & 99.45 & 59.87 \\
\multicolumn{1}{l}{} &
  GPT-3.5 Turbo &
   &
  36.83 &
  45.77 &
  99.83 &
  69.13 \\
\multicolumn{1}{l}{} &
  GPT-4 &
   &
  \textbf{37.65} &
  \textbf{47.02} &
  \textbf{99.37} &
  \textbf{70.75} \\ \cmidrule(l){2-7} 
\multicolumn{1}{l}{} &
  GPT-3.5 Davinci &
  \multirow{3}{*}{2-shot} &
  46.18 &
  49.12 &
  \textbf{69.0} &
  73.9 \\
\multicolumn{1}{l}{} &
  GPT-3.5 Turbo &
   &
  \textbf{45.95} &
  49.79 &
  74.53 &
  74.63 \\
\multicolumn{1}{l}{} &
  GPT-4 &
   &
  45.37 &
  \textbf{50.26} &
  79.29 &
  \textbf{74.9} \\ \bottomrule
\end{tabular}
}
\caption{Comparing GPT-3.5 \emph{text-davinci-003} to ChatGPT models \emph{gpt-3.5-turbo} and \emph{gpt-4} for zero-shot and few-shot translation with 2 fuzzy matches}
\label{tab:chatgpt}
\end{table}

%\vspace*{-14pt}

\section{BLOOM and BLOOMZ}

In this section, we compare GPT-3.5 to open-source multilingual models, namely BLOOM \cite{BLOOM2022} and BLOOMZ \cite{Muennighoff2022-BLOOMZ-mT0}. While BLOOM is a general-purpose LLM, BLOOMZ belongs to a family of models capable of following human instructions in a zero-shot manner.

\enlargethispage{0.5\baselineskip}

We use BLOOM and BLOOMZ via the Hugging Face's Inference API.\footnote{\url{https://huggingface.co/inference-api}} As mentioned in Section \ref{sec:setup}, recommended (sampling) parameters for translation with \mbox{GPT-3.5} are top-p 1 and temperature up to 0.3. For BLOOM, the same parameters are not good for translation.\footnote{Using lower sampling values of top-p and temperature such as 0.9 and 0.1, respectively, can generate good outputs. However, greedy search shows better translation performance.} We found that ``greedy search" achieves better results for BLOOM, which are reported in Table \ref{tab:bloom}. We use a batch size of 1, and set the $max\_new\_tokens$ parameter to be double the number of words of the source sentence if it is less than 250, the maximum number of new tokens allowed by BLOOM's API; otherwise, we set it to 250 tokens. For comparison purposes, we use the same values for BLOOMZ.\footnote{BLOOMZ is trained to generate the required output only; however, using BLOOM, we had to truncate over-generated text outputs, excluding anything generated in a new line.}

When providing each system with two fuzzy matches, generally GPT-3.5 outperforms both BLOOM and BLOOMZ for most language pairs, except English-to-Arabic translation. The English-to-French translation quality of BLOOM and \mbox{GPT-3.5} is comparable.

\begin{table}[H]
\captionsetup{font=scriptsize,labelfont=scriptsize}
\centering
\scalebox{0.56}{
\begin{tabular}{@{}clcccc@{}}
\toprule
\textbf{Lang}                   & \multicolumn{1}{c}{\textbf{System}} & \textbf{spBLEU ↑} & \textbf{chrF++ ↑} & \textbf{TER ↓} & \textbf{COMET ↑} \\ \midrule
\multirow{3}{*}{\textbf{EN-AR}} & BLOOM fuzzy 2-shot  & \textbf{43.19} & \textbf{59.48}   & \textbf{57.58} &                                  \textbf{67.36}   \\
                                & BLOOMZ fuzzy 2-shot & 36.29          & 53.33            & 66.86       &
                                58.4           \\
                                & GPT-3 fuzzy 2-shot & 38.41          & 56.57          & 62.31          & 57.36          \\ \cmidrule(l){2-6} 
\multirow{3}{*}{\textbf{EN-ES}} & BLOOM fuzzy 2-shot & 57.67          & 74.25          & 34.86          &                                       86.48          \\
                                & BLOOMZ fuzzy 2-shot & 53.07          & 70.44         & 40.45          & 81.38         \\
                                & GPT-3 fuzzy 2-shot  & \textbf{59.64} & \textbf{75.83} & \textbf{32.56} & \textbf{90.37} \\ \cmidrule(l){2-6} 
\multirow{3}{*}{\textbf{EN-FR}} & BLOOM fuzzy 2-shot  & \textbf{50.52} & 66.81         & \textbf{46.45} &                                     55.74            \\
                                & BLOOMZ fuzzy 2-shot & 45.1           & 62.73         & 51.69           &
                                47.49             \\
                                & GPT-3 fuzzy 2-shot & 49.79          & \textbf{67.41} & 46.79          & \textbf{61.38} \\ \cmidrule(l){2-6} 
\multirow{3}{*}{\textbf{EN-RW}} & BLOOM fuzzy 2-shot & 10.95          & 31.87          & 91.07 & N/A                                   \\
                                & BLOOMZ fuzzy 2-shot & \textbf{12.26}          & 35.44          & \textbf{88.36}          & N/A \\
                                & GPT-3 fuzzy 2-shot & 12.23          & \textbf{36.66} & 105.54         & N/A            \\ \cmidrule(l){2-6} 
\multirow{3}{*}{\textbf{EN-ZH}} & BLOOM fuzzy 2-shot & 40.62          & 40.62          & 75.24          &                                     66.23          \\
                                & BLOOMZ fuzzy 2-shot & 34.82         & 38.23          & 80.03          & 59.92          \\
                                & GPT-3 fuzzy 2-shot & \textbf{46.18} & \textbf{49.12} & \textbf{69.0}  & \textbf{73.9}  \\ \bottomrule
\end{tabular}
}
\caption{Comparing GPT-3.5 to BLOOM and BLOOMZ for few-shot translation with 2 fuzzy matches}
\label{tab:bloom}
\end{table}

\section{Conclusion}
%\vspace*{-0.51mm}

In this work, we conducted several experiments to assess the performance of GPT-3.5 across multiple translation tasks, namely adaptive MT using fuzzy matches (cf. Section \ref{sec:adaptive-MT}), MT post-editing (cf. Section \ref{sec:fix-mt}), terminology extraction \mbox{(cf. Section \ref{sec:term-extract})}, and terminology-constrained MT (cf. Section \ref{sec:term-constrain}). Moreover, we compared its translation quality with strong encoder-decoder MT systems. Generally speaking, results obtained from these experiments are very promising. While some high-resource languages such as English-to-French, English-to-Spanish and even English-to-Chinese show excellent results, other languages have lower support either because they are low-resource languages such as English-to-Kinyarwanda or because of \mbox{issues} in the \mbox{GPT-3.5} tokenizer such as English-to-Arabic. Nevertheless, when we used GPT-3.5 for MT post-editing of the English-to-Arabic translation obtained from OPUS, the quality significantly surpassed that obtained from both OPUS and Google Translation API. This means that different pipelines can be adopted in production for different language pairs, based on the level of support of these languages by an LLM.

Furthermore, we briefly compared \mbox{GPT-3.5} translation quality with open-source LLMs such as BLOOM and BLOOMZ. In the future, we would like to expand our experiments with open-source LLMs to cover more aspects.

For adaptive MT with fuzzy matches, it would be interesting to investigate \emph{dynamic} few-shot example selection. For instance, instead of selecting 5 fuzzy matches for all sentences, only high-quality fuzzy matches up to a certain similarity score are used. Similarly, when incorporating glossary terms or MT outputs from other systems, only those with certain quality characteristics are utilized. This can potentially enhance performance gains.

For terminology extraction, we would like to try ``phrases" instead of ``terms". This would generate longer strings. We would like to see the effect of using such longer phrases, especially for low-resource languages.

This work mainly aims at understanding the quality and level of support that LLMs can achieve (out of the box) for a range of translation tasks across diverse language pairs. In the future, we might consider starting with fine-tuning the model, and then conducting similar experiments. This can be especially beneficial for low-resource languages and rare domains, and can help enhance quality and efficiency.

\section*{Acknowledgements}
This work is supported by the Science Foundation Ireland (SFI) Centre for Research Training in Digitally-Enhanced Reality (d-real) under Grant No. 18/CRT/6224, the ADAPT Centre for \mbox{Digital} Content Technology under SFI's Grant No. 13/RC/2106\_P2, and Microsoft Research.

We would like to extend our sincere thanks to Julie Locquet, Senior Linguist; Philippe Locquet, Senior Linguist and Academic Program Manager at Wordfast; and Dr Muhammed Yaman Muhaisen, Ophthalmologist and Linguist, for conducting the evaluation of our translation tasks.

\footnotesize
\bibliography{paperpile}

\begin{thebibliography}{}

\bibitem[\protect\citename{Agrawal \bgroup et al.\egroup
  }2022]{Agrawal2022-SelectionMT}
Agrawal, Sweta, Chunting Zhou, Mike Lewis, Luke Zettlemoyer, and Marjan
  Ghazvininejad.
\newblock 2022.
\newblock {In-context Examples Selection for Machine Translation}.
\newblock {\em arXiv [cs.CL]}, December.

\bibitem[\protect\citename{Anastasopoulos \bgroup et al.\egroup
  }2020]{Anastasopoulos2020-TICO-19}
Anastasopoulos, Antonios, Alessandro Cattelan, Zi-Yi Dou, Marcello Federico,
  Christian Federmann, Dmitriy Genzel, Franscisco Guzm{\'a}n, et~al.
\newblock 2020.
\newblock {{TICO}-19: the Translation Initiative for {CO}vid-19}.
\newblock In {\em {Proceedings of the 1st Workshop on {NLP} for {COVID}-19
  (Part 2) at {EMNLP} 2020}}, Online, December.

\bibitem[\protect\citename{{BigScience Workshop} \bgroup et al.\egroup
  }2022]{BLOOM2022}
{BigScience Workshop}, Teven Le~Scao, Angela Fan, Christopher Akiki, Ellie
  Pavlick, Suzana Ili{\'c}, Daniel Hesslow, et~al.
\newblock 2022.
\newblock {BLOOM: A 176B-Parameter Open-Access Multilingual Language Model}.
\newblock {\em arXiv [cs.CL]}, November.

\bibitem[\protect\citename{Brown \bgroup et al.\egroup }2020]{Brown2020-GPT-3}
Brown, Tom~B, Benjamin Mann, Nick Ryder, Melanie Subbiah, Jared Kaplan,
  Prafulla Dhariwal, Arvind Neelakantan, et~al.
\newblock 2020.
\newblock {Language Models are Few-Shot Learners}.
\newblock In {\em {Advances in Neural Information Processing Systems (NeurIPS
  2020)}}, volume~33, pages 1877--1901.

\bibitem[\protect\citename{Bulte and Tezcan}2019]{Bulte2019-fuzzy}
Bulte, Bram and Arda Tezcan.
\newblock 2019.
\newblock {Neural Fuzzy Repair: Integrating Fuzzy Matches into Neural Machine
  Translation}.
\newblock In {\em {Proceedings of the 57th Annual Meeting of the Association
  for Computational Linguistics}}, pages 1800--1809, Florence, Italy, July.

\bibitem[\protect\citename{Chowdhery \bgroup et al.\egroup
  }2022]{Chowdhery2022-PaLM}
Chowdhery, Aakanksha, Sharan Narang, Jacob Devlin, Maarten Bosma, Gaurav
  Mishra, Adam Roberts, Paul Barham, et~al.
\newblock 2022.
\newblock {PaLM: Scaling Language Modeling with Pathways}.
\newblock {\em arXiv [cs.CL]}, April.

\bibitem[\protect\citename{Coughlin}2003]{Coughlin2003-MT-eval}
Coughlin, Deborah.
\newblock 2003.
\newblock {Correlating automated and human assessments of machine translation
  quality}.
\newblock In {\em {Proceedings of Machine Translation Summit IX: Papers}}, New
  Orleans, USA.

\bibitem[\protect\citename{Dinu \bgroup et al.\egroup
  }2019]{Dinu2019-TerminologyConstraintsTraining}
Dinu, Georgiana, Prashant Mathur, Marcello Federico, and Yaser Al-Onaizan.
\newblock 2019.
\newblock {Training Neural Machine Translation to Apply Terminology
  Constraints}.
\newblock In {\em {Proceedings of the 57th Annual Meeting of the Association
  for Computational Linguistics}}, pages 3063--3068, Florence, Italy, July.

\bibitem[\protect\citename{Dong \bgroup et al.\egroup
  }2022]{Dong2022-In-contextLearning}
Dong, Qingxiu, Lei Li, Damai Dai, Ce~Zheng, Zhiyong Wu, Baobao Chang, Xu~Sun,
  Jingjing Xu, Lei Li, and Zhifang Sui.
\newblock 2022.
\newblock {A Survey on In-context Learning}.
\newblock {\em arXiv [cs.CL]}, December.

\bibitem[\protect\citename{Etchegoyhen \bgroup et al.\egroup
  }2021]{Etchegoyhen2021-OnlineLearning}
Etchegoyhen, Thierry, David Ponce, Harritxu Gete, and Victor Ruiz.
\newblock 2021.
\newblock {Online Learning over Time in Adaptive Neural Machine Translation}.
\newblock In {\em {Proceedings of the International Conference on Recent
  Advances in Natural Language Processing (RANLP 2021)}}, pages 411--420, Held
  Online, September.

\bibitem[\protect\citename{Farajian \bgroup et al.\egroup
  }2017]{Farajian2017-AdaptiveMT}
Farajian, M~Amin, Marco Turchi, Matteo Negri, and Marcello Federico.
\newblock 2017.
\newblock {Multi-Domain Neural Machine Translation through Unsupervised
  Adaptation}.
\newblock In {\em {Proceedings of the Second Conference on Machine
  Translation}}, pages 127--137, Copenhagen, Denmark, September.

\bibitem[\protect\citename{Goyal \bgroup et al.\egroup }2022]{Goyal2022-spBLEU}
Goyal, Naman, Cynthia Gao, Vishrav Chaudhary, Peng-Jen Chen, Guillaume Wenzek,
  Da~Ju, Sanjana Krishnan, Marc'aurelio Ranzato, Francisco Guzm{\'a}n, and
  Angela Fan.
\newblock 2022.
\newblock {The Flores-101 evaluation benchmark for low-resource and
  multilingual machine translation}.
\newblock {\em Trans. Assoc. Comput. Linguist.}, 10:522--538, May.

\bibitem[\protect\citename{Haque \bgroup et al.\egroup
  }2020]{Haque2020-Terminology}
Haque, Rejwanul, Yasmin Moslem, and Andy Way.
\newblock 2020.
\newblock {Terminology-Aware Sentence Mining for {NMT} Domain Adaptation:
  {ADAPT}{'}s Submission to the Adap-{MT} 2020 {E}nglish-to-{H}indi {AI}
  Translation Shared Task}.
\newblock In {\em {Proceedings of the 17th International Conference on Natural
  Language Processing (ICON): Adap-MT 2020 Shared Task}}, pages 17--23, Patna,
  India, December.

\bibitem[\protect\citename{Hokamp and Liu}2017]{Hokamp2017-ConstrainedDecoding}
Hokamp, Chris and Qun Liu.
\newblock 2017.
\newblock {Lexically Constrained Decoding for Sequence Generation Using Grid
  Beam Search}.
\newblock In {\em {Proceedings of the 55th Annual Meeting of the Association
  for Computational Linguistics (Volume 1: Long Papers)}}, pages 1535--1546,
  Vancouver, Canada, July.

\bibitem[\protect\citename{Hosseini \bgroup et al.\egroup
  }2020]{Hosseini2020-DeepMatch}
Hosseini, Kasra, Federico Nanni, and Mariona Coll~Ardanuy.
\newblock 2020.
\newblock {{D}eezy{M}atch: A Flexible Deep Learning Approach to Fuzzy String
  Matching}.
\newblock In {\em {Proceedings of the 2020 Conference on Empirical Methods in
  Natural Language Processing: System Demonstrations}}, pages 62--69, Online,
  October.

\bibitem[\protect\citename{Hu \bgroup et al.\egroup
  }2019]{Hu2019-LexiconInduction}
Hu, Junjie, Mengzhou Xia, Graham Neubig, and Jaime Carbonell.
\newblock 2019.
\newblock {Domain Adaptation of Neural Machine Translation by Lexicon
  Induction}.
\newblock In {\em {Proceedings of the 57th Annual Meeting of the Association
  for Computational Linguistics}}, pages 2989--3001, Florence, Italy, July.

\bibitem[\protect\citename{Klein \bgroup et al.\egroup
  }2020]{Klein2020-Efficient}
Klein, Guillaume, Dakun Zhang, Cl{\'e}ment Chouteau, Josep Crego, and Jean
  Senellart.
\newblock 2020.
\newblock {Efficient and high-quality neural machine translation with
  {OpenNMT}}.
\newblock In {\em {Proceedings of the Fourth Workshop on Neural Generation and
  Translation}}, pages 211--217, Stroudsburg, PA, USA, July.

\bibitem[\protect\citename{Knowles \bgroup et al.\egroup
  }2018]{Knowles2018-Fuzzy}
Knowles, Rebecca, John Ortega, and Philipp Koehn.
\newblock 2018.
\newblock {A Comparison of Machine Translation Paradigms for Use in Black-Box
  Fuzzy-Match Repair}.
\newblock In {\em {Proceedings of the {AMTA} 2018 Workshop on Translation
  Quality Estimation and Automatic Post-Editing}}, pages 249--255, Boston, MA,
  March.

\bibitem[\protect\citename{Kudo and Richardson}2018]{Kudo2018-SentencePiece}
Kudo, Taku and John Richardson.
\newblock 2018.
\newblock {{S}entence{P}iece: A simple and language independent subword
  tokenizer and detokenizer for Neural Text Processing}.
\newblock In {\em {Proceedings of the 2018 Conference on Empirical Methods in
  Natural Language Processing: System Demonstrations}}, pages 66--71, Brussels,
  Belgium, November.

\bibitem[\protect\citename{Michon \bgroup et al.\egroup
  }2020]{Michon2020-Terminology}
Michon, Elise, Josep Crego, and Jean Senellart.
\newblock 2020.
\newblock {Integrating Domain Terminology into Neural Machine Translation}.
\newblock In {\em {Proceedings of the 28th International Conference on
  Computational Linguistics}}, pages 3925--3937, Barcelona, Spain (Online),
  December. International Committee on Computational Linguistics.

\bibitem[\protect\citename{Moslem \bgroup et al.\egroup
  }2022]{Moslem2022-MT-LM}
Moslem, Yasmin, Rejwanul Haque, John Kelleher, and Andy Way.
\newblock 2022.
\newblock {Domain-Specific Text Generation for Machine Translation}.
\newblock In {\em {Proceedings of the 15th biennial conference of the
  Association for Machine Translation in the Americas (Volume 1: Research
  Track)}}, pages 14--30, Orlando, USA, September.

\bibitem[\protect\citename{Muennighoff \bgroup et al.\egroup
  }2022]{Muennighoff2022-BLOOMZ-mT0}
Muennighoff, Niklas, Thomas Wang, Lintang Sutawika, Adam Roberts, Stella
  Biderman, Teven Le~Scao, M~Saiful~Bari, et~al.
\newblock 2022.
\newblock {Crosslingual Generalization through Multitask Finetuning}.
\newblock {\em arXiv [cs.CL]}, November.

\bibitem[\protect\citename{{NLLB Team} \bgroup et al.\egroup
  }2022]{NLLB_Team2022}
{NLLB Team}, Marta~R Costa-juss{\`a}, James Cross, Onur {\c C}elebi, Maha
  Elbayad, Kenneth Heafield, Kevin Heffernan, et~al.
\newblock 2022.
\newblock {No Language Left Behind: Scaling Human-Centered Machine
  Translation}.
\newblock {\em arXiv [cs.CL]}, July.

\bibitem[\protect\citename{Ouyang \bgroup et al.\egroup
  }2022]{Ouyang2022-InstructGPT}
Ouyang, Long, Jeff Wu, Xu~Jiang, Diogo Almeida, Carroll~L Wainwright, Pamela
  Mishkin, Chong Zhang, et~al.
\newblock 2022.
\newblock {Training language models to follow instructions with human
  feedback}.
\newblock {\em arXiv [cs.CL]}, March.

\bibitem[\protect\citename{Peris and
  Casacuberta}2019]{Peris2019-OnlineLearning}
Peris, {\'A}lvaro and Francisco Casacuberta.
\newblock 2019.
\newblock {Online learning for effort reduction in interactive neural machine
  translation}.
\newblock {\em Comput. Speech Lang.}, 58:98--126, November.

\bibitem[\protect\citename{Pham \bgroup et al.\egroup }2020]{Pham2020-Priming}
Pham, Minh~Quang, Jitao Xu, Josep Crego, Fran{\c c}ois Yvon, and Jean
  Senellart.
\newblock 2020.
\newblock {Priming Neural Machine Translation}.
\newblock In {\em {Proceedings of the Fifth Conference on Machine
  Translation}}, pages 516--527, Online, November.

\bibitem[\protect\citename{Post and
  Vilar}2018]{Post2018-FastConstrainedDecoding}
Post, Matt and David Vilar.
\newblock 2018.
\newblock {Fast Lexically Constrained Decoding with Dynamic Beam Allocation for
  Neural Machine Translation}.
\newblock In {\em {Proceedings of the 2018 Conference of the North {A}merican
  Chapter of the Association for Computational Linguistics: Human Language
  Technologies, Volume 1 (Long Papers)}}, pages 1314--1324, New Orleans,
  Louisiana, June.

\bibitem[\protect\citename{Reimers and
  Gurevych}2019]{Reimers2019-SentenceTransformers}
Reimers, Nils and Iryna Gurevych.
\newblock 2019.
\newblock {Sentence-{BERT}: Sentence Embeddings using {S}iamese
  {BERT}-Networks}.
\newblock In {\em {Proceedings of the 2019 Conference on Empirical Methods in
  Natural Language Processing and the 9th International Joint Conference on
  Natural Language Processing (EMNLP-IJCNLP)}}, pages 3982--3992, Hong Kong,
  China, November.

\bibitem[\protect\citename{Tiedemann}2020]{Tiedemann2020-Tatoeba}
Tiedemann, J{\"o}rg.
\newblock 2020.
\newblock {The Tatoeba Translation Challenge {--} Realistic Data Sets for Low
  Resource and Multilingual {MT}}.
\newblock In {\em {Proceedings of the Fifth Conference on Machine
  Translation}}, pages 1174--1182, Online, November.

\bibitem[\protect\citename{Touvron \bgroup et al.\egroup
  }2023]{Touvron2023-LLaMA}
Touvron, Hugo, Thibaut Lavril, Gautier Izacard, Xavier Martinet, Marie-Anne
  Lachaux, Timoth{\'e}e Lacroix, Baptiste Rozi{\`e}re, et~al.
\newblock 2023.
\newblock {LLaMA: Open and Efficient Foundation Language Models}.
\newblock {\em arXiv [cs.CL]}, February.

\bibitem[\protect\citename{Vaswani \bgroup et al.\egroup
  }2017]{Vaswani2017-attention}
Vaswani, Ashish, Noam Shazeer, Niki Parmar, Jakob Uszkoreit, Llion Jones,
  Aidan~N Gomez, Lukasz Kaiser, and Illia Polosukhin.
\newblock 2017.
\newblock {Attention Is All You Need}.
\newblock In {\em {Advances in Neural Information Processing Systems (NIPS
  2017)}}, volume~30.

\bibitem[\protect\citename{Vilar \bgroup et al.\egroup
  }2022]{Vilar2022-PaLM-Translation}
Vilar, David, Markus Freitag, Colin Cherry, Jiaming Luo, Viresh Ratnakar, and
  George Foster.
\newblock 2022.
\newblock {Prompting PaLM for Translation: Assessing Strategies and
  Performance}.
\newblock {\em arXiv [cs.CL]}, November.

\bibitem[\protect\citename{Wang \bgroup et al.\egroup }2021]{Wang2021-LM4MT}
Wang, Shuo, Zhaopeng Tu, Zhixing Tan, Wenxuan Wang, Maosong Sun, and Yang Liu.
\newblock 2021.
\newblock {Language Models are Good Translators}.
\newblock {\em ArXiv}.

\bibitem[\protect\citename{Wuebker \bgroup et al.\egroup
  }2018]{Wuebker2018-Personalized}
Wuebker, Joern, Patrick Simianer, and John DeNero.
\newblock 2018.
\newblock {Compact Personalized Models for Neural Machine Translation}.
\newblock In {\em {Proceedings of the 2018 Conference on Empirical Methods in
  Natural Language Processing}}, pages 881--886, Brussels, Belgium.

\bibitem[\protect\citename{Xu \bgroup et al.\egroup }2020]{Xu2020-fuzzy}
Xu, Jitao, Josep Crego, and Jean Senellart.
\newblock 2020.
\newblock {Boosting Neural Machine Translation with Similar Translations}.
\newblock In {\em {Proceedings of the 58th Annual Meeting of the Association
  for Computational Linguistics}}, pages 1580--1590, Online, July.

\bibitem[\protect\citename{Zhang \bgroup et al.\egroup
  }2023]{Zhang2023-PromptingMT}
Zhang, Biao, Barry Haddow, and Alexandra Birch.
\newblock 2023.
\newblock {Prompting Large Language Model for Machine Translation: A Case
  Study}.
\newblock {\em arXiv [cs.CL]}, January.

\end{thebibliography}
\bibliographystyle{eamt23}

% Number of pages: \thepage{}

\newpage
\appendix

\section{Prompts}
\label{a-prompts}
\vspace*{2.5pt}

This appendix provides examples of the prompts we used for our experiments.

\vspace*{7pt}
\subsection{Zero-shot Translation}
\vspace*{2pt}
% Zero-shot translation

\begin{tcolorbox}[enhanced,attach boxed title to top left={yshift=-3mm,yshifttext=-1mm,xshift=3mm},
  colback=blue!1!white,colframe=cyan!90!black,colbacktitle=cyan!90!black,boxrule=1pt,
  left=3pt, top=0pt, width=210pt,center,
  title=Prompt: EN-AR zero-shot translation,fonttitle=\small,
  boxed title style={size=small,colframe=cyan!90!black} ]
  \begin{scriptsize}
  \textmyfont{\emph{
    \begin{itemize}
    \setlength\itemsep{-0.8ex}
    \item[] English: $<$source\_segment$>$
    \item[] Arabic:
\end{itemize}
}}
\end{scriptsize}
\end{tcolorbox}

\vspace*{5pt}
\subsection{Adaptive MT with Fuzzy Matches}
\vspace*{2pt}
% Few-shot translation

\begin{tcolorbox}[enhanced,attach boxed title to top left={yshift=-3mm,yshifttext=-1mm,xshift=3mm},
  colback=blue!1!white,colframe=cyan!90!black,colbacktitle=cyan!90!black,boxrule=1pt,
  left=3pt, top=0pt, width=210pt,center,
  title=Prompt: EN-AR two-shot translation,fonttitle=\small,
  boxed title style={size=small,colframe=cyan!90!black} ]
  \begin{scriptsize}
  \textmyfont{\emph{
    \begin{itemize}
    \setlength\itemsep{-0.8ex}
    \item[] English: $<$source\_fuzzy\_match\textsubscript{2}$>$
    \item[] Arabic: $<$target\_fuzzy\_match\textsubscript{2}$>$
    \item[] English: $<$source\_fuzzy\_match\textsubscript{1}$>$
    \item[] Arabic: $<$target\_fuzzy\_match\textsubscript{1}$>$
    \item[] English: $<$source\_segment$>$
    \item[] Arabic:
\end{itemize}
}}
\end{scriptsize}
\end{tcolorbox}

\vspace*{5pt}
\subsection{MT Post-editing}
\vspace*{2pt}
% Two-shot + 1-MT

\begin{tcolorbox}[enhanced,attach boxed title to top left={yshift=-3mm,yshifttext=-1mm,xshift=3mm},
  colback=blue!1!white,colframe=cyan!90!black,colbacktitle=cyan!90!black,boxrule=1pt,
  left=3pt, top=0pt, width=210pt,center,
  title=Prompt: EN-ZH two-shot + 1-MT,fonttitle=\small,
  boxed title style={size=small,colframe=cyan!90!black} ]
  \begin{scriptsize}
  \textmyfont{\emph{
    \begin{itemize}
    \setlength\itemsep{-0.8ex}
    \item[] English: $<$source\_fuzzy\_match\textsubscript{2}$>$
    \item[] Chinese: $<$target\_fuzzy\_match\textsubscript{2}$>$
    \item[] English: $<$source\_fuzzy\_match\textsubscript{1}$>$
    \item[] Chinese: $<$target\_fuzzy\_match\textsubscript{1}$>$
    \item[] English: $<$source\_segment$>$
    \item[] MT: $<$mt\_segment$>$
    \item[] Chinese:
\end{itemize}
}}
\end{scriptsize}
\end{tcolorbox}

% Two-shot + all-MT
\vspace*{2pt}

\begin{tcolorbox}[enhanced,attach boxed title to top left={yshift=-3mm,yshifttext=-1mm,xshift=3mm},
  colback=blue!1!white,colframe=cyan!90!black,colbacktitle=cyan!90!black,boxrule=1pt,
  left=3pt, top=0pt, width=210pt,center,
  title=Prompt: EN-ZH two-shot + all-MT,fonttitle=\small,
  boxed title style={size=small,colframe=cyan!90!black} ]
  \begin{scriptsize}
  \textmyfont{\emph{
    \begin{itemize}
    \setlength\itemsep{-0.8ex}
    \item[] English: $<$source\_fuzzy\_match\textsubscript{2}$>$
    \item[] MT: $<$mt\_fuzzy\_match\textsubscript{2}$>$
    \item[] Chinese: $<$target\_fuzzy\_match\textsubscript{2}$>$
    \item[] English: $<$source\_fuzzy\_match\textsubscript{1}$>$
    \item[] MT: $<$mt\_fuzzy\_match\textsubscript{1}$>$
    \item[] Chinese: $<$target\_fuzzy\_match\textsubscript{1}$>$
    \item[] English: $<$source\_segment$>$
    \item[] MT: $<$mt\_segment$>$
    \item[] Chinese:
\end{itemize}
}}
\end{scriptsize}
\end{tcolorbox}

\newpage
\subsection{Terminology Extraction}
\vspace*{2pt}
% Terminology extraction

\begin{tcolorbox}[enhanced,attach boxed title to top left={yshift=-3mm,yshifttext=-1mm,xshift=3mm},
  colback=blue!1!white,colframe=cyan!90!black,colbacktitle=cyan!90!black,boxrule=1pt,
  left=-3pt, top=0pt, width=210pt,center,
  title=Prompt: terminology extraction,fonttitle=\small,
  boxed title style={size=small,colframe=cyan!90!black} ]
  \begin{scriptsize}
  \textmyfont{\emph{
    \begin{itemize}
    \setlength\itemsep{-0.8ex}
    \item[] $<$source\_lang$>$: $<$source\_sentence$>$
    \item[] $<$target\_lang$>$: $<$target\_sentence$>$
    \item[]
    \item[] Extract $<$number$>$ terms from the above sentence pair. Type each $<$source\_lang$>$ term and its $<$target\_lang$>$ equivalent in one line, separated by '$<$separator$>$'.
    \item[]
    \item[] 1.
\end{itemize}
}}
\end{scriptsize}
\end{tcolorbox}

\vspace*{5pt}
\subsection{Terminology-constrained MT}
\vspace*{2pt}
% Zero-shot + glossary terms

\begin{tcolorbox}[enhanced,attach boxed title to top left={yshift=-3mm,yshifttext=-1mm,xshift=3mm},
  colback=blue!1!white,colframe=cyan!90!black,colbacktitle=cyan!90!black,boxrule=1pt,
  title=Prompt: EN-ES zero-shot + glossary terms,fonttitle=\small,
  left=-3pt, top=0pt, width=210pt,center,
  boxed title style={size=small,colframe=cyan!90!black} ]
  \begin{scriptsize}
  \textmyfont{\emph{
  \begin{itemize}
    \setlength\itemsep{-0.8ex}
    \item[] Terms: $<$src\_term\textsubscript{1}$>$ $=$ $<$tgt\_term\textsubscript{1}$>$ - $<$src\_term\textsubscript{2}$>$ $=$ $<$tgt\_term\textsubscript{2}$>$ ... $<$src\_term\textsubscript{5}$>$ $=$ $<$tgt\_term\textsubscript{5}$>$
    \item[] English: $<$source\_segment$>$
    \item[] Spanish:
\end{itemize}
}}
\end{scriptsize}
\end{tcolorbox}

\vspace*{2pt}
% two-shot + fuzzy terms

\begin{tcolorbox}[enhanced,attach boxed title to top left={yshift=-3mm,yshifttext=-1mm,xshift=3mm},
  colback=blue!1!white,colframe=cyan!90!black,colbacktitle=cyan!90!black,boxrule=1pt,
  title=Prompt: EN-ES two-shot + fuzzy terms,fonttitle=\small,
  left=3pt, top=0pt, width=210pt,center,
  boxed title style={size=small,colframe=cyan!90!black} ]
  \begin{scriptsize}
  \textmyfont{\emph{
  \begin{itemize}
    \setlength\itemsep{-0.8ex}
    \item[] Terms: $<$terms\_fuzzy\_match\textsubscript{2}$>$
    \item[] English: $<$source\_fuzzy\_match\textsubscript{2}$>$
    \item[] Spanish: $<$target\_fuzzy\_match\textsubscript{2}$>$
    \item[] Terms: $<$terms\_fuzzy\_match\textsubscript{1}$>$
    \item[] English: $<$source\_fuzzy\_match\textsubscript{1}$>$
    \item[] Spanish: $<$target\_fuzzy\_match\textsubscript{1}$>$
    \item[] Terms: $<$terms\_from\_fuzzy\_matches\textsubscript{1+2}$>$
    \item[] English: $<$source\_segment$>$
    \item[] Spanish:
\end{itemize}
}}
\end{scriptsize}
\end{tcolorbox}

\vspace*{2pt}
% Two-shot + glossary terms

\begin{tcolorbox}[enhanced,attach boxed title to top left={yshift=-3mm,yshifttext=-1mm,xshift=3mm},
  colback=blue!1!white,colframe=cyan!90!black,colbacktitle=cyan!90!black,boxrule=1pt,
  title=Prompt: EN-ES two-shot + glossary terms,fonttitle=\small,
  left=3pt, top=0pt, width=210pt,center,
  boxed title style={size=small,colframe=cyan!90!black} ]
  \begin{scriptsize}
  \textmyfont{\emph{
  \begin{itemize}
\setlength\itemsep{-0.8ex}
    \item[] Terms: $<$terms\_fuzzy\_match\textsubscript{2}$>$
    \item[] English: $<$source\_fuzzy\_match\textsubscript{2}$>$
    \item[] Spanish: $<$target\_fuzzy\_match\textsubscript{2}$>$
    \item[] Terms: $<$terms\_fuzzy\_match\textsubscript{1}$>$
    \item[] English: $<$source\_fuzzy\_match\textsubscript{1}$>$
    \item[] Spanish: $<$target\_fuzzy\_match\textsubscript{1}$>$
    \item[] Terms: $<$terms\_from\_glossary$>$
    \item[] English: $<$source\_segment$>$
    \item[] Spanish:
\end{itemize}
}}
\end{scriptsize}
\end{tcolorbox}

\end{document}